\documentclass[letterpaper, 10 pt, conference]{ieeeconf}

\IEEEoverridecommandlockouts                              

\usepackage{graphicx}
\usepackage{caption}
\usepackage{subcaption}
\usepackage{amsmath,amssymb}
\usepackage[bookmarks=true]{hyperref}

\usepackage{algorithm}
\usepackage{algorithmic}
\usepackage[algo2e, ruled]{algorithm2e}

\newcount\Comments 
\usepackage{array,multirow,graphicx}
\usepackage{color}
\usepackage{wrapfig}
\definecolor{darkgreen}{rgb}{0,0.5,0}
\definecolor{purple}{rgb}{1,0,1}
\newcommand{\kibitz}[2]{\ifnum\Comments=1\textcolor{#1}{#2}\fi}

\DeclareMathOperator*{\argmin}{arg\,min}
\DeclareMathOperator*{\argmax}{arg\,max}

\title{\LARGE \bf
Accelerated Robot Learning via Human Brain Signals
}

\author{Iretiayo Akinola\thanks{$^*$Equal Contribution}$^{*1}$, Zizhao Wang$^{*1}$, Junyao Shi$^{1}$, Xiaomin He$^{2}$, Pawan Lapborisuth$^{2}$,\\ Jingxi Xu$^{1}$, David Watkins-Valls$^{1}$, Paul Sajda$^{2,3}$ and Peter Allen$^{1}$
\thanks{This work was supported in part by a Google Research grant and National Science Foundation grant IIS-1527747.
} 
\thanks{$^{1}$Department of Computer Science, Columbia University, New York}%
\thanks{$^{2}$Department of Biomedical Engineering, Columbia University, New York}%
\thanks{$^{3}$Data Science Institute, Columbia University, New York, NY 10027}%
}

\begin{document}

\maketitle
\thispagestyle{empty}
\pagestyle{empty}

\begin{abstract}
In reinforcement learning (RL), sparse rewards are a natural way to specify the task to be learned. However, most RL algorithms struggle to learn in this setting since the learning signal is mostly zeros.
In contrast, humans are good at assessing and predicting the future consequences of actions and can serve as good reward/policy shapers to accelerate the robot learning process.
Previous works have shown that the human brain generates an error-related signal, measurable using electroencephelography (EEG), when the human perceives the task being done erroneously.
In this work, we propose a method that uses evaluative feedback obtained from human brain signals measured via scalp EEG to accelerate RL for robotic agents in sparse reward settings. As the robot learns the task, the EEG of a human observer watching the robot attempts is recorded and decoded into noisy error feedback signal.
From this feedback, we use supervised learning to obtain a policy that subsequently augments the behavior policy and guides exploration in the early stages of RL.
This bootstraps the RL learning process to enable learning from sparse reward.
Using a simple robotic navigation task as a test bed, we show that our method achieves a stable obstacle-avoidance policy with high success rate, outperforming learning from sparse rewards only that struggles to achieve obstacle avoidance behavior or fails to advance to the goal.
\end{abstract}


\section{Introduction}
Reinforcement Learning (RL) remains one of the most popular learning approaches because of its simplicity and similarity to how humans learn from reward signals. Also, it achieves superior performance on a number of robotic tasks. However, RL requires defining a good reward function that captures the task to be learned, and deriving an appropriate reward function remains a challenge. The sparse reward is a natural way to specify a task; here the agent receives a positive feedback only when the task has been accomplished and nothing otherwise. This sparse reward formulation is easy to set up, and when it works, it is unlikely to produce unusual artifact behavior due to local optima. A drawback is that it provides poor learning signals especially when the task horizon is long. Since RL learns by trial and error, the chances that the agent would accidentally achieve the task's goal is very small in the sparse reward setting. This makes RL from sparse rewards very challenging or sometimes impossible. 
A few methods have been devised to address this problem. For example, reward shaping is a common approach of designing rich reward functions that can better guide the learning process\cite{mataric1994reward}\cite{ng1999policy}. Reward function design can be a laborious iterative process requiring expert knowledge and some art. Alternatively, non-expert demonstrations can be used to initialize and augment the learning process \cite{atkeson1997robot}\cite{vevcerik2017leveraging}\cite{nair2018overcoming}. This is a simple and effective method. However, it requires that the task is first demonstrated by a human, which is not always possible.

\begin{figure}[t]
\vspace{2mm}
\begin{center}
    \begin{subfigure}[h]{1\linewidth}
        \centering
        \includegraphics[width=0.75\linewidth]{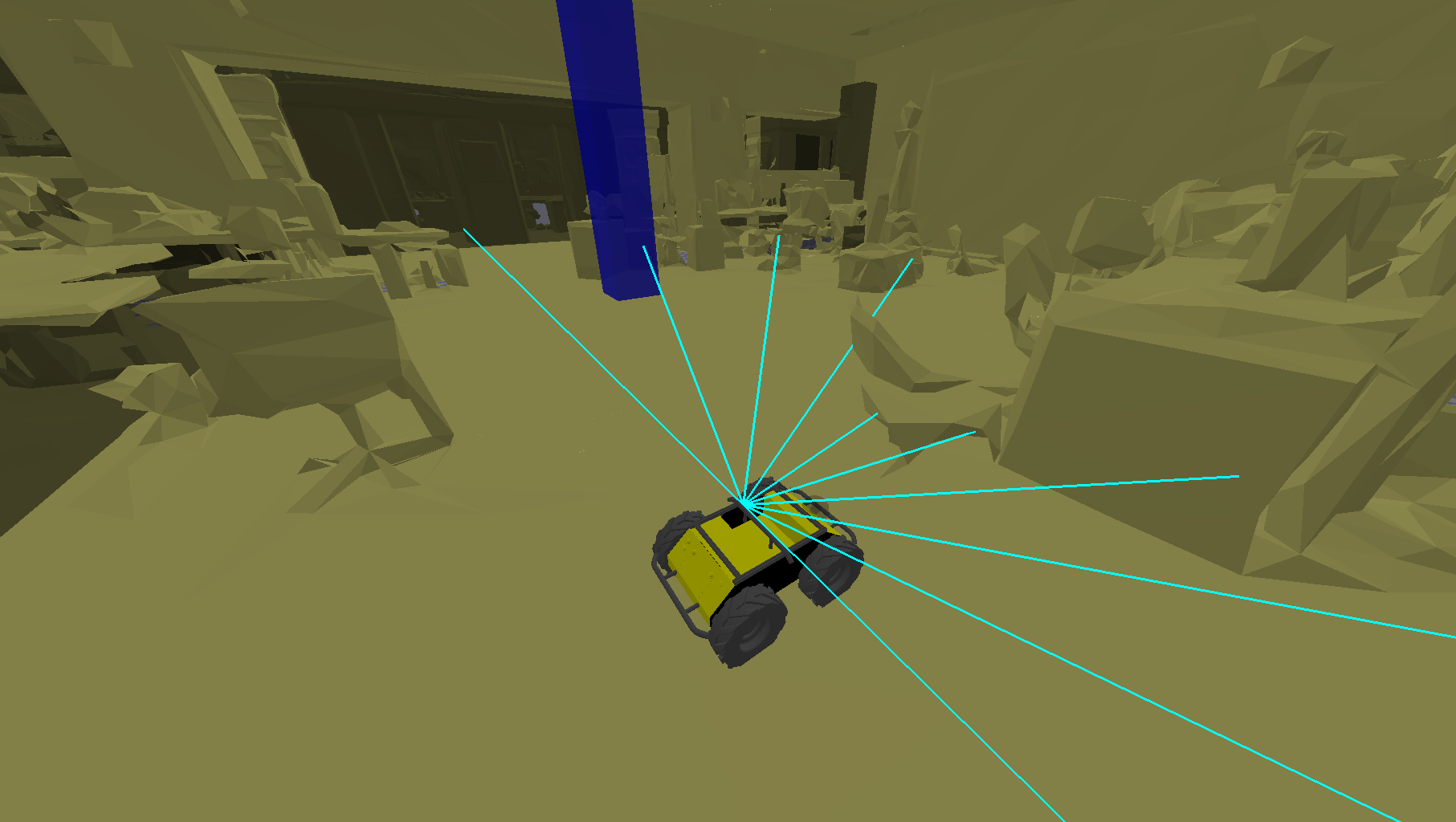}
    \end{subfigure}
    \par\smallskip 
    \begin{subfigure}[h]{1\linewidth}
        \centering
        \includegraphics[width=0.75\linewidth]{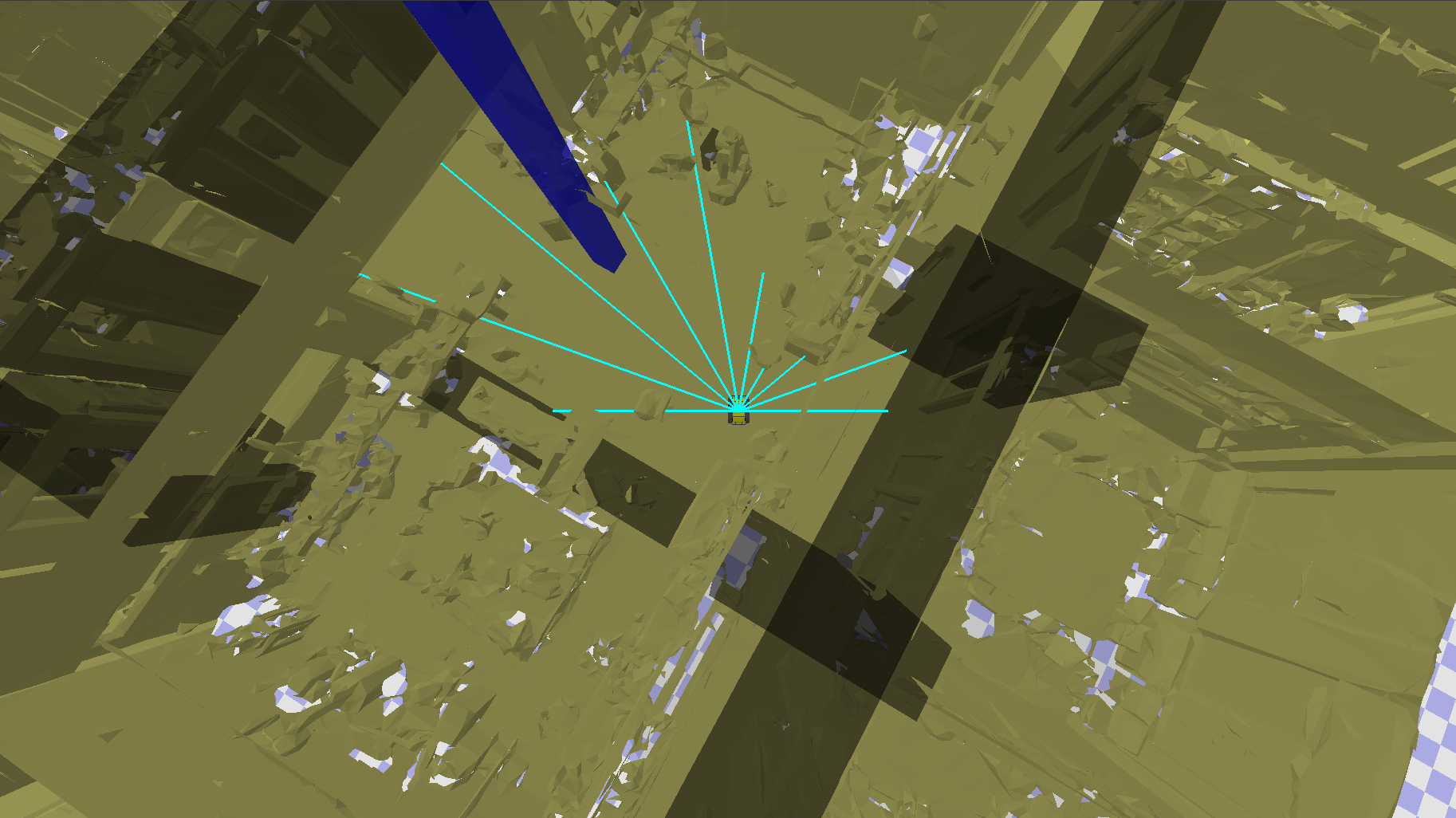}
    \end{subfigure}
\end{center}
\setlength{\abovecaptionskip}{-0pt}
\setlength{\belowcaptionskip}{-21pt}
\caption{\small Navigation Task. The robot agent, given its current orientation, its goal location and an ability to sense its environment with laser scans (indicated by the 10 light cyan rays), learns to navigate to a goal (blue) without colliding with obstacles. In the sparse reward RL setting, the agent is unable to avoid obstacles and reach the goal. A third-person view (Top) is shown to the subjects during training and our method of using human EEG as evaluative feedback to accelerate the early learning phase enables the agent to learn this navigation task.}
\label{fig:nav_task}
\end{figure}

Another class of methods has humans providing feedback to the agent as it learns.
Learning from human feedback is an increasingly popular approach to teaching robots different skills \cite{knox2009interactively}\cite{griffith2013policy}\cite{christiano2017deep}\cite{warnell2018deep}. One reason is that this approach resembles how humans learn from instructor feedback, as in a school setting. Another reason is that learning from feedback fits into the reinforcement learning paradigm where the feedback signal can be used as the reward signal. It can also be used in the supervised learning setting where actions are classified as good or bad at a given state; the agent learns to take actions classified as good. Since humans tend to have a general idea of how certain tasks should be done, and are quite good at predicting the future consequences of actions, feedback from human experts provides a natural and useful signal to train artificial agents such as robots.
In this work, we adopt the learning from feedback approach, where the feedback is the error signal detected from the brain of a human watching the agent learn.
Previous work in neuroscience has shown that a distinctive error signal such as error-related potential (ErrP) occurs in the human brain when the human observes an error during a task \cite{spuler2015error}. We exploit this ability of a human expert to pick out erroneous actions committed by an apprentice robot during training.

Learning directly from human brain activity is appealing for a number of reasons. It presents a convenient way to transfer human knowledge of the tasks into an artificial agent, even when it is difficult to provide precise, explicit instructions. For tasks that can be easily assessed by a human, evaluative feedback is detected with little latency since the human does not need to react by pressing a button or other input, thus  providing a temporally-local credit assignment. However, there are a few problems that need to be addressed: detecting ErrP signals with sufficient accuracy to be useful during the early stages of learning, keeping the user engaged during observations, and reducing the amount of human feedback needed for the learning process.

In this work, we examine key issues around learning from human brain signals and seek answers to a number of questions, including the following:
\begin{itemize}
  \item How can artificial agents learn directly from human physiological signals, such as brain signals?
  \item What is a good way to combine learning from human brain signals with task-success sparse reward signals?
  \item How does the learning performance change with the error signal detection accuracy?
\end{itemize}

To answer these questions, we first simulate the ErrP-based feedback signals using a noisy oracle. This oracle detects whether an agent's action was the optimal action and gives the feedback accordingly. Using different oracle accuracy levels, we are able to do extensive analysis on the behaviour of different task learning algorithms. Ultimately, we monitor their performance when the accuracy of the oracle feedback is set at a level that matches that of the human brain signal classifier. 
Based on the extensive simulated analysis, we obtain a robust algorithm that can learn from noisy human feedback such as human brain signals. Second, we demonstrate this in physical experiments where EEG signals from human subjects are used to improve an agent's learning of a navigation task in a sparse reward setting (see fig.\ref{fig:nav_task}).  
On multiple navigation tasks, our Brain-Guided RL outperforms learning from baselines using sparse or rich rewards. It also shows robustness to low ErrP detection accuracy.

\begin{figure*}[ht]
\vspace{2mm}
\begin{center}
    \includegraphics[width=0.85\linewidth]{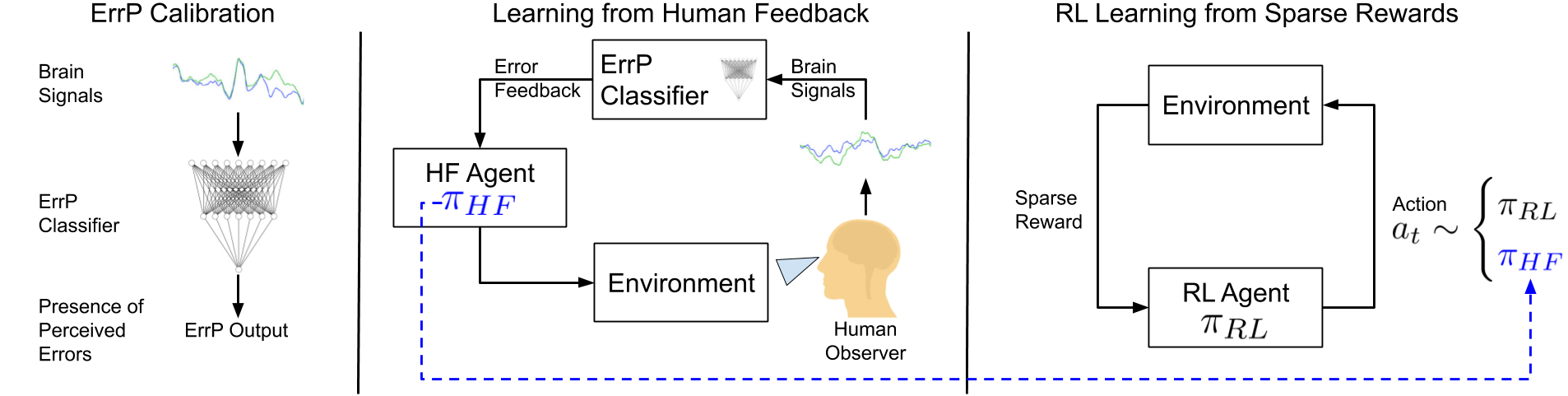}
\end{center}
\setlength{\belowcaptionskip}{-15pt}
\caption{\small Brain-Guided RL in three stages. \textbf{Left}: The ErrP calibration stage where a function is learned to detect error potential from human brain signal. \textbf{Middle}: A human observer watches an agent learn a task and evaluative feedback is tapped from the human brain and provided to the agent. A policy $\pi_\text{HF}$ is learned to choose actions that avoid negative human feedback. \textbf{Right}: The RL agent learns from sparse rewards but the behavior policy during the learning process is a blend of the RL policy ($\pi_\text{RL}$) and the human feedback policy ($\pi_\text{HF}$). $\pi_\text{HF}$ (learned in stage two) helps guide the exploration so that the agent sees more positive rewards required for RL learning.
}
\label{fig:bci_robot_learning_stages}
\end{figure*}

\section{Background and Related Work}

\subsection{Reinforcement Learning}
Reinforcement learning (RL) is an area of machine learning concerned with how agents act to maximize cumulative reward from the environment. The reward captures the objective of the task such that the cumulative reward is maximized when the task is achieved.
Mathematically, an RL problem can be formulated as a Markov Decision Process (MDP) that consists of a set of states $S$, a set of actions $A$, a transition function $T : S \times A \rightarrow \Delta S$, and a reward function $R : S \times A \rightarrow \Delta \mathbb{R}$. A policy $\pi : S \rightarrow A$ is learned to maximize the cumulative reward $\sum_{t=0}^{\infty} \gamma^t R(s_t, a_t)$, where $0 \le \gamma \le 1$ is the discount factor.
The reward function can be discontinuous with sparse rewards given at milestones or it can be complex and continuous to capture progress toward the goal.
Specifying a good reward function is key to defining an RL problem. This generally entails using domain knowledge to define a function that captures progress on the tasks in terms of the state and action spaces. Such informative reward functions can be complicated and difficult to realize, requiring a laborious iterative process.

\subsection{Learning from Feedback}
An alternative approach to defining the reward function is getting feedback from an expert who evaluates the action (or sequence of actions) taken by the agent during an episode and provides a score. This score can serve as the reward in the RL framework.
Previous works used different interfaces to collect humans feedback. Some collect binary feedback via mouse clicks \cite{christiano2017deep}\cite{jagodnik2017training} or a more graduated feedback \cite{jagodnik2017training} via sliders. Others utilize facial expressions \cite{veeriah2016face}, and finger pointing \cite{cruz2016multi} among others.
In our work, we obtain the feedback information directly from the human brain.
\cite{iturrate2015teaching} demonstrated that brain signals can be used to learn control policies for navigating in one- and two- dimensional environments; these discrete state spaces are relatively small, with a total of 8 and 13 distinct locations respectively. The size is the constrained because expert labels are expensive to obtain. In our work, we present a way to analyze the learning behavior in larger state spaces and how it varies with the error rate of the feedback.

\subsection{The ErrP}
Previous works have covered a few types of ErrP, including the Response, Observation, Feedback, and Interaction ErrPs \cite{dias2018masked}. The \textit{response} ErrP occurs when a subject makes an error while responding to a stimulus within a short amount of time; the \textit{observation} ErrP occurs when observing another agent make a mistake; the \textit{feedback}  ErrP  occurs  when a subject  receive a negative assessment of the subject's action; and the \textit{interaction} ErrP occurs when the subject senses a mismatch between the subject's command and the interface's response. While the paradigms are different, similar signal processing and machine learning techniques are used to detect the different types of ErrP. In this work, we are interested in the \textit{observation} ErrP as evaluative feedback to robot agents during learning.

To calibrate such a detector, the EEG signals of human subjects are recorded and time-locked to error onsets. These signals then go through several pre-processing steps that include filtering, artifact removal, and subsampling. A classifier is then trained on the processed signal to differentiate brain activity when an error is being observed. The classification performance reported in the literature ranges from slightly above chance to 0.8 \cite{welke2017brain} \cite{ehrlich2018feasibility}.
A recent work \cite{welke2017brain} examined brain activities during robot-error observations, and their findings indicated relatively low decoding accuracies of observation ErrPs compared to other ErrP types. They concluded that further improvements  in  non-invasive recording and analysis techniques are necessary for practical applications.
In this work, we develop a method to utilize the observation ErrP as a complement to learning from sparse rewards despite the low ErrP decoding accuracy.

\subsection{Brain-Computer Interface (BCI) Robot Learning}
BCI has been used in robotics to issue commands that directly control robots~\cite{bi2013eeg}\cite{choi2013low}\cite{zhang2016control}\cite{akinola2017task},  correct robot mistakes~\cite{salazar2017correcting}, and guide the robot to goals inferred from the brain signals~\cite{iturrate2013shared}.
In these works, the brain error signals prompt the robots to change the current course of action, but they do not result in an autonomous skill that persists when the human is no longer observing.
Recent work \cite{schiatti2018human} has used similar ErrPs as reward signals for teaching a behavior to the robot so the robot can autonomously achieve the task after training.
Our work differs from these existing works in that:
\begin{itemize}
    \item we address the well-known issue of the rarity for the sparse rewards setting by leveraging noisy human brain signals to guide exploration and accelerate the early stage of learning.
    \item we do not require that the human subject be involved in the entire training cycle.  Human feedback is expensive to obtain and our method shows that only limited human feedback may be needed.
    \item we retain the ability to do reinforcement learning via easily specifiable sparse reward signals and achieve good-quality asymptotic performance on the task.
    \item our formulation ensures that the learning process is not limited by the low signal-to-noise ratio of BCI signals.
    \item we demonstrate the applicability of our algorithm to realistic autonomous mobile navigation-- an important research area in robotics.
\end{itemize}

\section{Method}

Our Brain-Guided RL algorithm works in three stages (See Figure \ref{fig:bci_robot_learning_stages}): train a classifier on EEG signals to detect occurrences of human-perceived error, learn a Human Feedback (HF) policy using the trained EEG classifier, and learn the final RL policy from sparse rewards as the HF policy guides RL exploration.
In the first stage, we collect EEG signals, the robot actions and corresponding ground truth correct actions. We infer the human feedback label to be an \textit{error} whenever the robot action does not match ground truth. For example, if the robot turned left but the correct action is to turn right, we assign an error label to that move. The recorded brain signals and the feedback labels are used to train the EEG classifier offline to detect ErrPs.
For the second stage, a human subject watches the robot agent take actions on the target task and concurrently we apply the trained classifier on the brain signals to detect the human's feedback online. Based on this feedback, a supervised learning model is trained online to predict the probability that an action gets a positive feedback.
The robot's policy is continuously updated by maximizing this success probability across possible actions -- we refer to this as the HF policy.
Lastly, an agent is trained on the same task with RL from sparse rewards, guided by the HF policy to improve exploration. 

\subsection{EEG Classifier Training}

To obtain evaluative feedback from the human brain, we need a function that maps EEG brain signals to ErrP labels (correct/incorrect) for the observed robot actions. This is done during a calibration stage where we collect data offline to train an EEG classifier. In the data collection step, the human subject watches an agent conducting a random policy while we simultaneously record EEG signals and the labels indicating if actions are correct or erroneous. 
The robot takes an action every 1.5s so that the brain signals elicited by each action can be time-locked without interfering with subsequent actions. This slow speed also enables the human to assess each action in a way that elicits the strongest brain signals.
We use a navigation task for our analysis; here a user watches a mobile robot navigate to a target location. Wrong actions that move the robot away from the target or into obstacles will elicit responses in the subject's brain.
Using the Dijkstra search algorithm, we obtain the optimal action at each step which provides ground-truth labels for good versus bad actions. A human expert can also provide these ground truth labels, especially for tasks whose optimal solutions cannot be easily scripted.
In our experiments, the EEG signals are recorded at 2048 Hz using 64 channels of the BioSemi EEG Headset and around 600 data points of robot actions are collected.

After data collection, we preprocess the EEG data and train the classifier. During preprocessing, the data is band-pass filtered to 1-40 Hz to remove artificial noise and resampled to 128 Hz. EEG trials are extracted at [0, 0.8]s post the agent action. Then, each processed EEG data $x_i$ around a robot action $a_i$ is used as input for the classifier to predict the corresponding label $f_i$. 
Our classifier, denoted as $g(\cdot; \theta_\text{EEG})$, is modified from EEGNet \cite{lawhern2018eegnet}\footnote{For convolution layers of the EEGNet, we change to valid padding and reduce the number of filters $(F1=4, D=2, F2=4)$ to alleviate overfitting.}. EEGNet is a compact network with temporal and depthwise convolutions to capture frequency-based spatial features.
$80 \%$ of the data are used for training, while $12 \%$ and $8 \%$ are held out for validation and testing respectively. We optimize the classifier with the cross-entropy loss $\mathcal{L}_\text{EEG}$.

\begin{equation}\label{key}
\theta_{EEG}^* = \argmin_{\theta_{EEG}} \frac{1}{M} \sum_{i=1}^M [\mathcal{L}_{EEG}(g(x_i; \theta_{EEG}), f_i)]
\end{equation}

After training, the classifier $g(x_t; \theta_\text{EEG}^*)$ maps the EEG signal $x_t$ to human feedback $f_t$ as the subject observes the agent executing an action $a_t$, indicating if the action was erroneous or not.
The testing accuracy ranges from $55 \%$ to $75 \%$, depending on the subjects, which means large noise exists in the feedback.

\subsection{Human Feedback Policy}

With the EEG classifier from the previous section, we can tell (from the human brain) if an observed action is correct or not.
Instead of directly using this human feedback as a reward function for RL as in some previous work \cite{iturrate2010robot} \cite{schiatti2018human}, we use it in a supervised learning setting to learn the human feedback function $F$ for the target task.
This target task may be different from the task used in the EEG calibration step.
The calibration task can be simpler (e.g navigation in a smaller room) where human feedback as ground truth labels is less expensive to collect.
Formally, when the agent executes the action $a_t$ at the state $s_t$, the human observes and judges whether $a_t$ is the optimal action captured by $F(s_t, a_t)$. Using the classified brain signal $f_t$ as noisy labels, we learn an approximation of $F$ which we denote as $\hat{F}$ and construct an HF policy from it given as:
\begin{equation}
\pi_\text{HF}(s) = \argmax_{a}\hat{F}(s, a)
\end{equation}

Learning $\hat{F}$ is exactly supervised learning: the input is the agent experience $(s_t, a_t)$ and the label is the human feedback $f_t$.
$\hat{F}$ can be any function approximator; we use a fully-connected neural network in this work.
This function is learned in an online fashion; $\hat{F}$ is continually updated with data as the robot acts based on the $\pi_\text{HF}$.
The challenges here are: the limited amount of human feedback (1000 labels) and the inconsistent label $f_t$ due to the noise from the EEG classifier. To mitigate this, we adopt three strategies: (1) reduce the number of parameters by choosing low-dimensional continuous state and action spaces (2) design a light network architecture (3) use a feedback replay buffer.
We use a fully-connected network with 1 hidden layer of 16 units and one output node for each action. The predicted optimality for a state-action pair, $\hat{F}(s_t, a_t)$, is obtained by passing $s_t$ as input to the network and select the output node corresponding to $a_t$. During training, we use the cross entropy loss and only backpropagate through the single output node for the observed action.
We keep $20 \%$ of feedback as validation data to confirm that there is no clear overfitting.
To learn the parameters quickly, the network is updated at a faster rate than the rate of human feedback by reusing feedback labels.
We adopt a feedback replay buffer which is a priority queue that stores all agent experiences $(s_t, a_t)$ and the corresponding human feedback $f_t$; newer experiences are of greater importance. Batches of data are continually pulled from the replay buffer to optimize the network $\hat{F}$.

At the end of the session, the policy $\pi_\text{HF}$ has a general notion of which actions are good/bad and how to perform the task. 
Although imperfect classification of noisy EEG signals limits the
performance of $\pi_\text{HF}$, it still provides better
exploration when doing RL in a sparse reward setting.

\subsection{Efficient Sparse-Reward RL with Guided Exploration}

The final stage is to enable the RL agent to learn efficiently in an environment with sparse rewards. The challenge here is that random exploration is unlikely to stumble on positive rewards that aids learning. To address this, we use $\pi_\text{HF}$ as the initial behavior policy during RL learning. Even though $\pi_\text{HF}$ may be far from perfect, this guides the exploration towards the goal and increase the chances of getting positive rewards. As learning proceeds, we reduce the use of $\pi_\text{HF}$ and increasingly use the learned RL policy as the behavior policy.
Eventually, the agent is able to learn the task as specified by the sparse reward function. Our full algorthim for Brain-Guided RL is given in Algorithm \ref{alg:brain_guided_rl}.

Implementation-wise, we can choose any off-policy Deep RL algorithm as the RL policy. Our method is even robust to on-policy Deep RL algorithms like PPO \cite{schulman2017proximal} which we adopt as the RL policy for the experiments. At the beginning of each episode, there is an $\epsilon_\text{HF}$ chance to use the HF policy for this episode. $\epsilon_\text{HF}$ linearly decays from $\epsilon_\text{HF, init}$ to 0 in the first $t_\text{trans}$ time steps. After the RL policy learns the environment setting in the transition stage, the training is fully on-policy. The RL policy refines itself, gets beyond the suboptimal HF policy, and learns the optimal behavior.

\section{Experiments}
We use robot navigation tasks as the test-bed for our algorithm.
The tasks are implemented in the Gibson simulation environment \cite{xiazamirhe2018gibsonenv} as shown in Fig \ref{fig:nav_task}.
The Gibson environment is a high-fidelity simulation engine created from real world data of 1400 floor spaces from 572 full buildings. It models real-world's semantic complexity and enforces constraints of physics and space; it can detect collision and respects non-interpenetrability of rigid body, making it suitable for simulating navigation tasks in a realistic way.
We use a $11 \times 12 m^2$ area with multiple obstacles, and choose the Husky robot for our tasks.
The goal location is represented by the blue square pillar. In all navigation tasks, the position of the goal is fixed, since it is very challenging to learn a HF policy for a variable goal task within the limited amount of feedback (1000 labels).

\begin{algorithm2e}
\SetAlgoNoLine
\caption{\strut Brain-Guided RL}
\KwData{offline EEG signals $x_{1:M}$ and labels $f_{1:M}$, HF policy update epoch number $K_\text{HF}$, RL policy update epoch number $K_\text{RL}$}
\textbf{Train the EEG classifier.}
\begin{equation*}
\theta_\text{EEG}^* = \argmin_{\theta_\text{EEG}} \frac{1}{M} \sum_{i=1}^M [\mathcal{L}_\text{EEG}(g(x_i; \theta_\text{EEG}), f_i)].
\end{equation*}

\textbf{Train the HF policy.} \\
$B$ = [] $\#$ initialize the feedback replay buffer. \\
\For{$t = 1, 2, \dots, t_\text{HF}$}{
    observe state $s_t$. \\
    execute action $a_t = \pi_\text{HF}(s_t)$. \\
    receive human feedback by classifying EEG signal $f_t = g(x_t; \theta_\text{EEG}^*)$. \\
    update $\hat{F}$ using SGD with $((s_t, a_t), f_t)$. \\
    update $\hat{F}$ using SGD $K_\text{HF}$ epochs with minibatches sampled from $B$. \\
    append $((s_t, a_t), f_t)$ to $B$.
}
\textbf{Train the RL policy.} \\
\For{episode $i = 1, 2, \dots$}{
	$\epsilon_\text{HF} = \max(0, \epsilon_\text{HF,init} \cdot (1 - \frac{i \cdot T}{t_\text{trans}}))$. \\
	$\pi = \pi_\text{HF}$ with chance $\epsilon_\text{HF}$, otherwise $\pi = \pi_\text{RL}$. \\
	run policy $\pi$ for $T$ timesteps. \\ 
	optimize $\mathcal{L}_{PPO}$ using SGD $K_\text{RL}$ epochs with minibatches sampled from the episode. \\
}
\label{alg:brain_guided_rl}
\end{algorithm2e}


The state space is chosen as $s_t = (l_t, d_t, \phi_t) \in \mathbb{R}^{13}$ where $l_t \in \mathbb{R}^{10}$ is laser range observations evenly spaced between $-90^{\circ}$ and $90^{\circ}$ relative to the robot's frame, $d_t \in \mathbb{R}^{2}$ is displacement to the goal in global polar coordinates, and $\phi_t$ is the yaw of the robot.
The action space $A$ is discretized, as it is easier for the human subject to identify the actions and judge its optimality. We consider three actions: moving forward $0.3m$, turning $30^\circ$ left and turning $30^\circ$ right.

The task is to navigate from a start location to the goal without colliding with obstacles. This task can be captured by the sparse reward function \textbf{RL sparse} given as:
\[
    r_{sparse}(s_t, a_t) = 
\begin{cases}
    +100,  & \text{if reaches the goal} \\
    -100,  & \text{if collides with obstacles} \\
    -1,    & \text{otherwise}
\end{cases}
\]
Alternatively, we can design a richer, more-expressive reward function \textbf{RL rich} as:
\[
    r_{rich}(s_t, a_t) = r_{sparse}(s_t, a_t) + c_{d} \cdot d_t + c_\theta \cdot \theta_t 
\]
where $d_t$ is the euclidean distance from the goal, $\theta_t$ is the difference between the current orientation and the orientation to the goal, $c_d = -1.0$ and $c_\theta = -0.3$ are hyperparameters. This rich reward motivates the robot to get closer to and face the goal, leading to more efficient exploration and learning.
In the environment, we check if the agent reaches the goal through distance threshold checking ($0.5 m$). Reaching the goal or colliding with obstacles will end the episode.

\begin{figure}
\vspace{2mm}
\begin{center}
    \includegraphics[width=1.0\linewidth]{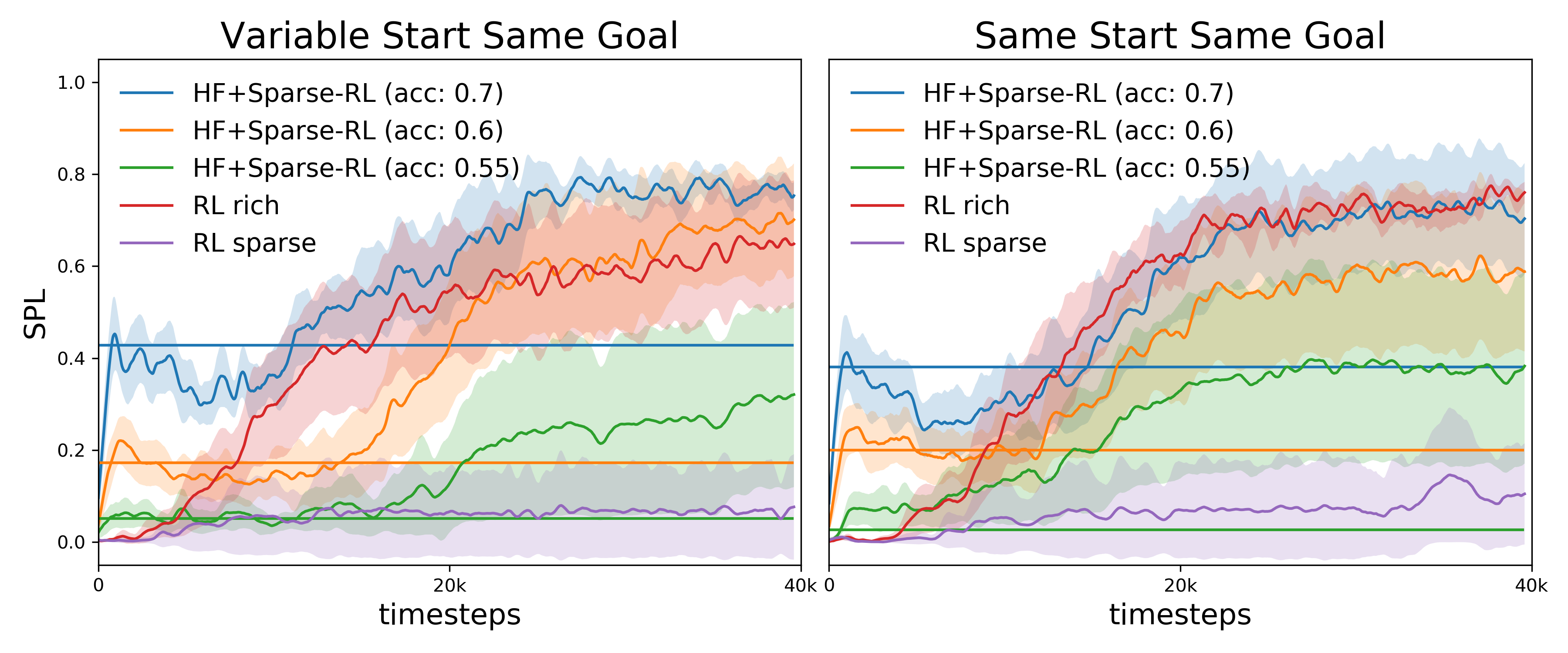}
\end{center}
\setlength{\belowcaptionskip}{-15pt}
\caption{\small Simulated Feedback Results. \textbf{Left}: Same Start Same Goal (\textbf{SSSG}), \textbf{Right}: Variable Start Same Goal (\textbf{VSSG}).
Using the SPL metric in both cases, we compare the performance of our method (HF+Sparse-RL) at varying feedback accuracy (Green: $70\%$, Orange: $60\%$, Blue: $55\%$) with RL-sparse (Purple) and RL-rich (Red). The plots show the mean and 1/2 of the standard deviation over 10 different runs. The horizontal lines represent the average performance of the learned HF policy at the corresponding feedback accuracy. When the feedback accuracy is $\ge 60\%$, feedback signals can be used to effectively accelerate reinforcement learning in sparse reward settings comparable to learning from a rich reward function.
Without guidance for feedback policy, learning from sparse reward is unable to learn.
}
\label{fig:sim_results}
\end{figure}

\begin{figure*}[h]
\vspace{2mm}
\begin{center}
    \includegraphics[width=1.0\linewidth]{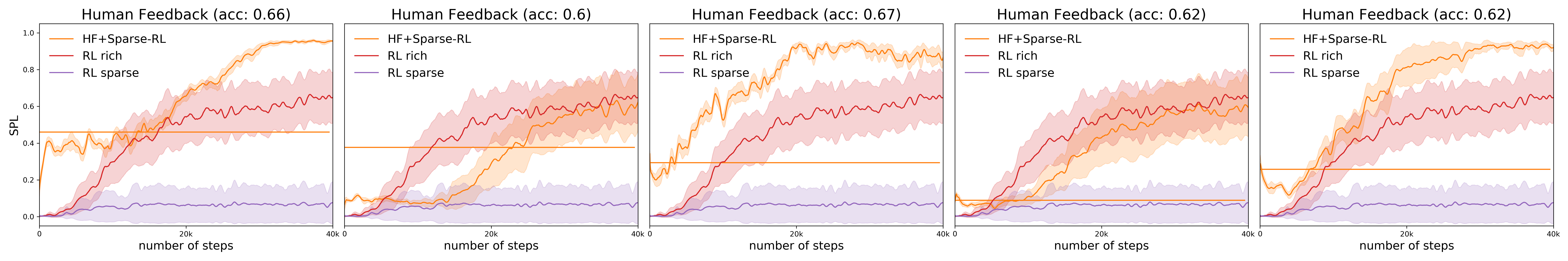}
\end{center}
\setlength{\belowcaptionskip}{-15pt}
\caption{\small Real EEG Feedback Results for 5 Successful Subjects. Our method (Orange) leverages feedback obtained from human brain signals (ErrPs) to accelerate the RL learning process and achieves superior asymptotic performance. HF policy ($\pi_\text{HF}$) is learned from a single online session and its performance is shown as the orange horizontal line. Afterwards, $\pi_\text{HF}$ guides RL policy learning in 5 different runs which are averaged out and shown in orange(HF+SparseRL).
For subjects (1 $\&$ 3) with higher ErrP detection accuracies, we observe bigger benefits from our method both in performance and learning speed. This is consistent with simulation results.}
\label{fig:real_results}
\end{figure*}

\section{Results}
In this section we evaluate our proposed algorithm on two variants of the navigation tasks: \textit{Same Start Same Goal} (\textbf{SSSG}) and \textit{Variable Start Same Goal} (\textbf{VSSG}).
For VSSG, the robot's starting location is uniformly chosen within a $0.2m \times 0.2m$ area. The optimal path takes 17 - 19 steps, while an episode is ended after 120 steps.
Beyond the scope of this work, this formulation can be extended to start the robot at any location by further expanding the starting square using curriculum learning.
To ensure repeatability and enable extensive analysis, we first use a simulated oracle to provide noisy feedback on the agent's actions. Then, we evaluate the performance of our system with human subjects using feedback from their EEG signals. For both simulation and real experiments, we report results comparing RL sparse, RL rich and HF+Sparse-RL (Ours). To assess the performance of all three methods, we adopt Success weighted by (normalized inverse) Path Length \cite{anderson2018evaluation} (SPL) which captures both success rate and path optimality. For fair comparison, we use the same architecture and hyperparameters for the RL part across all three methods.

\subsection{Learning from Simulated Feedback}

In the simulated setting, we vary the accuracy $\mathcal{C}\in \{0.55, 0.6, 0.7\}$ of the feedback coming from the simulated oracle and evaluate how well the HF policy assists the RL learning with noisy feedback.
Figure \ref{fig:sim_results} shows the result and $\mathcal{C} = 0.6$ matches the typical classification accuracy of the EEG classifier. Using grid search, we select the value of $\epsilon_\text{HF,init} = 0.8$ which decays linearly to $0$ after 50\% of the total training steps.
For both \textbf{SSSG} and \textbf{VSSG}, note that the sparse reward struggles to learn the task as it is rare to randomly stumble on the goal and obtain positive rewards required for learning. Our method (HF+Sparse-RL) solves the navigation task by using $\pi_\text{HF}$ policy obtained from noisy brain signals to guide the exploration and helps overcome the sparsity of the positive reward. The carefully-designed rich reward is also able to solve the navigation task but there are tasks where designing a rich reward function is prohibitively difficult. Our approach alleviates the need for such expert-level reward design process by combining evaluative human feedback and an easily specified sparse reward function.

\subsection{Learning from Real Human Feedback}

We tested our HF+Sparse-RL method on the VSSG task with 7 human subjects providing feedback in the form of EEG signals. First, the subject is trained for 5 minutes to get familiar with the paradigm and understand how to navigate the robot to the goal.
Then, the subject has a 20-min offline session to collect data for training the EEG classifier, a 5-min session to test classifier accuracy, and a 25-min online session to provide feedback and train the $\pi_\text{HF}$ policy.
This human feedback policy is subsequently used to guide the RL similar to the simulation experiments.
Video of the experiments can be found at \url{http://crlab.cs.columbia.edu/brain_guided_rl/}.

Shown in Figure \ref{fig:real_results}, the $\pi_\text{HF}$ policies from 5 subjects, with ErrP detection accuracy between 0.60 and 0.67, were able to successfully guide the learning process during RL from sparse reward. The EEG classifiers obtained for the other two subjects (accuracy of 0.56 and 0.57) were not good enough to train a useful $\pi_\text{HF}$ policy and thus could not guide RL learning.
Potential reasons include the subject not being engaged enough by the task to elicit ErrP or neurophysiological variations \cite{thompson2019critiquing} across subjects.

\section{Discussion}
The experiments on navigation tasks with feedback from either a simulated oracle and real humans show that Brain-Guided RL can accelerate RL in sparse reward environments.
Using human feedback directly as reward for RL seems appealing but it would require the human's attention for the entire training time which is typically very long for most RL algorithms.
Rather than directly applying feedback to RL learning, our Brain-Guided RL approach learns a HF policy via supervised learning in a relatively short session and then uses the learned policy to guide the RL agent. This choice saves a huge amount of expensive human feedback. 
It is also robust to low ErrP classification accuracy as a suboptimal HF policy can still improve RL exploration while allowing pure RL to achieve optimal performance.
Due to the low signal-to-noise ratio of the EEG device and a limited amount of human feedback, we were able to demonstrate our approach on a simple navigation task with little variance in the start/goal locations. To address this, we could use other approaches to further increase feedback efficiency; for example \textit{active learning} \cite{agarwal2013selective}\cite{orabona2011better}\cite{dekel2006online} can be used to determine which labels to query the user for. We leave this as a part of the future work. As BCI technology improves to achieve a higher signal-to-noise ratio, our approach can better scale to harder tasks. Our proposed method presents a potential for assistive robots to quickly learn new skills using inputs from humans with disabilities.

\section{Conclusion}

This paper introduces Brain-Guided RL, a method to accelerate RL learning in sparse reward settings, by using evaluative human feedback extracted from EEG brain signals. Our approach of first training a HF policy using supervised learning and then using it to guide RL learning demonstrates robustness in three important ways. It is robust to inconsistent feedback as is the case with noisy EEG signals and the resulting poor classification accuracy. It is also robust to the low performance of the policy obtained via the noisy human feedback since it still provides coarse guidance for the RL learning process. Finally, our approach reduces the the amount of feedback needed since the subject is not required to evaluate the robot's actions throughout the RL training process.
Experiments using both simulated and real human feedback show that our Brain-Guided RL enables learning different versions of the navigation task from sparse rewards with high success rate.
Future work includes using active learning techniques to maximize human feedback during the learning duration and adapting the proposed method to tasks with larger/continuous action spaces.

\section*{ACKNOWLEDGMENT}
We thank Carlos Martin for early versions of the simulated experimental analysis. We also thank Bohan Wu for valuable discussions when developing the algorithm, and everyone at Columbia Robotics Lab for useful comments and suggestions.
We gratefully acknowledge Microsoft Inc. for their support of Iretiayo Akinola through the Microsoft Research PhD Fellowship Program.


\bibliographystyle{IEEEtran}


\end{document}